\documentclass[letterpaper]{article} % DO NOT CHANGE THIS
\usepackage[]{aaai23}  % DO NOT CHANGE THIS
\usepackage{times}  % DO NOT CHANGE THIS
\usepackage{helvet}  % DO NOT CHANGE THIS
\usepackage{courier}  % DO NOT CHANGE THIS
\usepackage[hyphens]{url}  % DO NOT CHANGE THIS
\usepackage{graphicx} % DO NOT CHANGE THIS
\usepackage{natbib}  % DO NOT CHANGE THIS AND DO NOT ADD ANY OPTIONS TO IT
\usepackage{caption} % DO NOT CHANGE THIS AND DO NOT ADD ANY OPTIONS TO IT
\usepackage{amsmath}
\usepackage{amsfonts}
\usepackage{mathtools}
\usepackage{soul}
\usepackage{algorithm}
\usepackage{algorithmic}
\usepackage{subcaption}
\usepackage{enumitem}
\usepackage{cleveref}
\usepackage[table]{xcolor}
\usepackage{color}

\usepackage{newfloat}
\usepackage{listings}
\DeclareCaptionStyle{ruled}{labelfont=normalfont,labelsep=colon,strut=off} % DO NOT CHANGE THIS
\floatstyle{ruled}
\newfloat{listing}{tb}{lst}{}
\floatname{listing}{Listing}
\usepackage{color}

\title{Modeling Mobile Health Users as Reinforcement Learning Agents}
\author{
    Eura Shin,\textsuperscript{\rm 1}
    Siddharth Swaroop, \textsuperscript{\rm 1}
    Weiwei Pan, \textsuperscript{\rm 1}
    Susan Murphy, \textsuperscript{\rm 1}
    Finale Doshi-Velez, \textsuperscript{\rm 1}
}
\affiliations{
    \textsuperscript{\rm 1} Harvard University\\
}

\begin{document}

\maketitle

\begin{abstract}
Mobile health (mHealth) technologies empower patients to adopt/maintain healthy behaviors in their daily lives, by providing interventions (e.g.  push notifications) tailored to the user's needs. 
In these settings, without intervention, human decision making may be impaired (e.g. valuing near term pleasure over own long term goals). 
In this work, we formalize this relationship with a framework in which the user optimizes a (potentially impaired) Markov Decision Process (MDP) and the mHealth agent intervenes on the user's MDP parameters. We show that different types of impairments imply different types of optimal intervention. We also provide analytical and empirical explorations of these differences.
\end{abstract}

\section{Introduction}
In digital health, a human user and a mobile health (mHealth) application work together to acheive user specified behavioral goals. For example, the user may own a physical therapy (PT) app that guides them through a physician-recommended daily exercise routine. To plan effective intervention, the app agent maintains a model of the user's behavior. In our paper, the app agent models the user's decision making process as that of a reinforcement learning (RL) agent. As a result, there are \emph{two} RL agents that operate in this scenario. The first RL agent is the autonomous app agent, whose policy provides personalized interventions (e.g. a push notification about the importance of exercising daily) to the user in order to maintain healthy behavior (e.g. the PT routine). The second RL agent is the user, whose action space is binary -- they choose whether or not to engage in the suggested behavior (e.g. do the PT routine).  

Even though the user and app agents share the same goal -- long-term behavior change -- without intervention, the user's default decision may not be to engage in the target behavior, due to systematic impairments in the human's decision making. For example, the user may be myopic -- that is, they may heavily discounts future rewards \citep{story2014discountingAndHealthyBehavoir}. In our PT example, a fully rehabilitated shoulder may seem too distantly located in the future to motivate the user towards the goal. Prior work has tried to infer the user's impairment from demonstration. In this paper, we explore a complementary problem: assuming we know the user's impairment, what should the app agent do about it?

\textbf{Our contributions.} 
In this work, we explore effective ways for the app gent to intervene to maintain users' goal-oriented progress, in situations where the user would have otherwise disengaged. To do this, we propose a formal framework that represents the user as an RL agent, wherein different parametrizations of the user's \emph{Markov Decision Process} (MDP) capture a range of commonly observed user behaviors in mHealth -- state-dependent motivation, disengagement, and difficulty of adherence. For example, a myopic user is represented as an agent planning with a small discount factor $\gamma$. Furthermore, our framework formalizes the mechanism through which the app agent's interventions affect the user decisions; namely, the app agent intervenes on the user's MDP parameters. For example, intervening on a myopic user corresponds to increasing the user's discount factor $\gamma$. Finally, as a precursor to user studies, we use our framework to extract concrete intervention strategies that are expected to work well for a given type of user. We end with a discussion of the interesting behavioral and computational open questions that arise within this framework.

\textbf{Related work.} Reinforcement learning is frequently used to model the complex mechanisms underlying human behavior-- from the firing of dopaminergic neurons in the brain \citep{niv2009RLBrainDopamine, shteingart2014RLAndHumanBehavior} to disorders in computational psychiatry \citep{maia2011RLtoPsychiatricNeurological, chen2015RLinDepression}. In digital health, RL has been used to model maladaptive eating behaviors \citep{taylor2021MindfulnessEatingBehaviorRL}. Although these settings use RL to produce models of human decision making, the models themselves are not used to enrich planning for an autonomous agent. 

In some settings, such as in Human Robot Interaction \citep{tejwani2022socialRecursiveMDP, xie2021learningLatentMultiAgent, losey2019robotsAdvantageHumanTrust} or assistive AI \cite{chen2022mirrorAssistiveDriving, reddy2018InferringDynamics}, the human is modeled to inform the decisions of another RL algorithm. Though these applications can be formalized as a multi-agent RL problem (i.e. a two-player cooperative game), we formalize this as a single-agent problem, where we must solve for the mHealth agent's policy. This choice is motivated by our setting, where the strongest detectable effect of the app's intervention on the user tends to be immediate and transitory. 
%Thus  we model this setting as a single agent RL problem. 
%check out  Klasnja, P., Smith, S., Seewald, N.J., Lee, A.,Hall, K., Luers, B., Hekler, E.B. and Murphy, S.A. Efficacy of contextually-tailored suggestions for physical activity: A micro-randomized optimization trial of HeartSteps Ann Behav Med. 2019 Jun; 53(6): 573–582. and potentially cite

Modeling the actions of an external, human RL agent requires inferring the parameters that drive their policy. In Inverse Reinfocement Learning (IRL), this means inferring the user agent's \emph{reward function} \citep{ziebart2008MCEIRL}. IRL has been applied in digital health to infer user preferences-- \citet{zhou2018personalizingFitnessGoalRL} infer the user's utility function in order to set adaptive daily step goals. Similar to \citet{herman2016SERD} and \citet{reddy2018InferringDynamics}'s goals of inferring the transition dynamics, we are interested in modeling the user's entire decision making process, beyond the rewards. 

Despite evidence of humans demonstrating systematic behavioral impairments (e.g. myopic planning), most IRL approaches assume that humans agents act optimally, or near-optimally, with respect to a task. To improve on collaboration with humans, prior work has focused on representing and inferring these impairments from human actions \citep{evans2016learningIgnorant, laidlaw2022boltzmannPolicyDistribution, shah2019feasibilityOfLearning, jarrett2021inverseDecisionModeling}. However, the goal of prior work has been for the app agent to function in these collaborative settings \emph{despite} these impairments but not to \emph{intervene} on them.

\section{Computational Framework for Behavior Change in Digital Health}
\begin{figure}[t]
    \centering
    \includegraphics[width=1\linewidth]{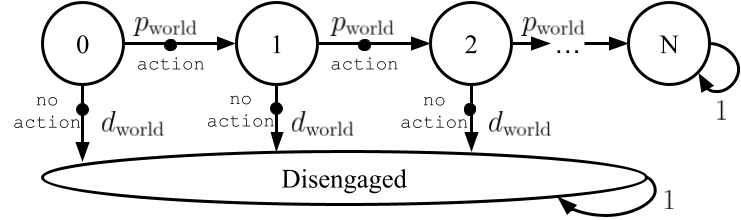}
    \caption{\textbf{Visualization of states and transitions in digital health gridworld}. Arrows are marked with the required action and probability of transitioning between states.}
    \label{fig: gridworld}
\end{figure}

\newcommand{\p}{{p_\text{user}}}
\newcommand{\dreal}{{d_\text{world}}}
\newcommand{\G}{{G_\text{user}}}
\newcommand{\B}{{B_\text{user}}}
\newcommand{\D}{{D_\text{user}}}
\newcommand{\gammau}{{\gamma_\text{user}}}
In this section, we capture the basic dynamics of a digital health application in a one-dimensional gridworld (visualized in \cref{fig: gridworld}): this is the environment in which the user and app agents operate.
Although simple, this environment captures many basic elements of behavior change in digital health, such as the difficulty of making progress, potential for disengagement, and goals.

The single dimension represents the user's current progress toward the behavior goal, for example, the current strength of the rehabilitated joint. Let $w = 1, \ldots, N$ represent this world state. There are two absorbing states: one for disengagement ($w = \text{disengaged}$), which happens when the user stops doing PT and/or deletes the app, and another for reaching a ``goal" ($w = N$), when the joint meets the desired level of strength. When the user practices PT, they go right with probability $p_\text{world} \in (0, 1]$, which means they increase the strength of the joint. If (and only if) the user abstains from PT, there is a chance that they disengage with probability $d_\text{world} \in (0, 1]$. 
%Note that if the user performs PT, there is no chance that they will disengage. 
For now it is impossible for the user to lose progress (go backwards). %

The two RL entities, the app agent and user agent, interface with this world in different ways. The user \emph{directly} interacts with the world-- they ultimately decide when and how to move between states (e.g. the app cannot perform the PT exercises for the user). The app \emph{indirectly} interacts with the world through the user; by influencing the user's decision making process, the app influences how the user moves in the world. In the next two sections, we formalize how the two agents perceive and plan actions within this world. 

\begin{table*}[t]
  \rowcolors{2}{gray!20}{white}
    \centering
    \begin{tabular}{| l | p{2.3cm} | p{13.1cm} | }
    \hline
         Interv. & Effect & Example \\
         \hline
         $a_B$ & Increases $\B$ by $\Delta_B$ & Use of implementation intentions \citep{wieber2015promotingImplementationIntention} to reduce the effort of starting the PT. \\ 
         $a_D$ & Decreases $\D$ by $\Delta_D$ & Use of multi-day ``streaks"; disengaging means losing the streak.\\
         $a_\gamma$ & Increases $\gammau$ by $\Delta_\gamma$ &  \citet{magen2008hiddenZero}'s hidden-zero framing: ``Skip PT today and lose shoulder mobility in the future OR do PT today and regain shoulder mobility in the future."\\ 
         $a_p$ & Increases $\p$ by $\Delta_p$ & Prompt self-monitoring of progress through graphs and metrics. \\
    \hline
    \end{tabular}
    \caption{\textbf{Definitions of app interventions.} The magnitude of $\Delta_\p$ can be interpreted as the effectiveness of the app's intervention on $\p$. We assume the $\Delta$'s are fixed over time. Examples are purely demonstrative and not final interventions.}
    \label{tab: app_interventions}
\end{table*}

\subsection{User agent: Model of user's decision making}
In this section we model the user's internal decision making process. The user's decision making is guided by a policy. The policy is \emph{optimal} for a Markov Decision Process $\mathcal{M}_\theta$, where $\theta$ is a set of user specific parameters defined below:

\noindent \textbf{State.} The user observes the app's intervention, which we call $a_\text{app}$ and the current progress towards the goal, $w$. 
%Only $w$ will be important to understanding the user's decisions; including $a_\text{app}$ is a technicality.  
%\sam{if short on space, consider deleting prior sentence}

\noindent \textbf{Action.} At each time step, the user decides to perform ($a_\text{user} = 1$) or not perform ($a_\text{user} = 0$) the behavior. For example, the user decides daily whether or not to do PT. 
%In our setting, we will want to encourage the user to take $a_\text{user} = 1$ more often. \sam{if short on space, consider deleting prior sentence}

\noindent \textbf{Rewards.} The user anticipates actions in certain states will have consequences. Doing PT may incur burden ($\B$) which the user perceives as a negative consequence. Reaching the goal state-- rehabilitating the shoulder-- may be of great positive consequence ($\G$). Finally, disengaging could either have a positive or negative consequence ($\D$). The user's perceived rewards are:  
\begin{equation}
\label{eq: user_reward}
 r_\mathrm{user}(w, a) = \begin{cases}
        \B, & \mathrm{for\ } a = 1\\
        \G, & \mathrm{for\ } w = N\\
        \D, & \mathrm{for\ } w = \text{Disengaged}\\
                 \end{cases}     
\end{equation}

%The form of the reward in \cref{eq: user_reward} is capable of expressing a range of attitudes that the user may have toward the behavior. A user that finds PT extremely unpleasant may have a large negative value for $\B$, but a user that enjoys PT may have a positive value.  

\noindent \textbf{Transitions.} 
%In making decisions, the user anticipates how their actions affect future states. 
%When the user does not perform the behavior, there is a probability of disengagement, $d_\text{world}$. 
The user believes that there is $p_\text{user}$ probability that doing the behavior will make them progress toward the goal. Note that there can be a difference between the user's \emph{perceived} and \emph{true} probability of making progress, $p_\text{world}$. For example, the user may think there is a low probability, $p_\text{user} = 0.1$, that doing PT will make them stronger. In actuality, the PT may be very effective, and the user progresses toward the goal at a much higher rate, $p_\text{world} = 0.7$. In this model, only the value of $\p$ affects the user's decision on whether or not to take an action. 

\noindent \textbf{Discount.} The user exponentially discounts future rewards via $\gammau \in [0, 1]$. 

\paragraph{}
To recap, the user's behavior is governed by the following parameters: $\theta = \{\B, \G, \D, \p, d_\text{real}, \gammau\}$. Learning these parameters corresponds to learning the user's decision making process. In practice, all of these parameters are internal to the user and must be inferred.

\subsection{App agent: digital intervention policy}
In this section, we define the app agent, who helps the user achieve their goal by maximizing the frequency at which the user performs the behavior. 

\textbf{State.} The app observes the world state $w$ and the user's $\hat a_\text{user}$. Unlike $a_\text{user}$, which is the user's \emph{intended} action,
%--- the action outputted by the user's optimal policy--- 
$\hat a_\text{user}$ is the \emph{observed} action. This accounts for some degree of randomness in the user's life: although they had planned to start PT for the day ($a_\text{user} = 1$), the user may receive an urgent phone call that prevents it from happening ($\hat a_\text{user} = 0$).
%--- the intended action corrupted by random noise, $\sigma^2$.
If the user decides $a_\text{user} = 1$, then the app observes $\hat a_\text{user} = 0$ with probability $\sigma^2$ and $a_\text{user} = 1$ with probability $1 - \sigma^2$. If the user decides $a_\text{user} = 0$, then $\hat a_\text{user} = 0$ always.  

\textbf{Action.} The action space consists of a separate intervention for each user parameter,  described in \cref{tab: app_interventions}.
%: $a_\text{app} \in \{a_p, a_\gamma, a_B, a_D\}$.  These actions along with their effects are.  
We exclude an intervention on $G$ because it is similar to an intervention on $\gamma$ (both increase motivation toward long term goal). We assume app actions affect the user's decision making \emph{in the moment}, and not permanently. For example, an intervention may increase $\gammau$ in one timestep, but it will revert to the original value in the next timestep, when the user has to decide again whether to perform the action. 

\textbf{Rewards.} The app receives a negative reward if the user takes no action and a positive reward if the user takes an action following intervention: $r_\text{app}(s_\text{app}, a_\text{app}) = 2 \hat a_\text{user} - 1$. 
%This definition of reward optimizes for the user's \emph{adherence} to the behavior. Another option would be to optimize for the user's \emph{performance} by making the reward depend on the user's distance from the goal state. However, in our formulation the user cannot move backwards, and maximizing either quantity would result in the same app policy, which makes the user move right as quickly as possible. 

\textbf{Transitions.} The app experiences transitions according to the \emph{true} probability of making progress ($p_\text{world}$ instead of $p_\text{user}$) and the \emph{true} probability that the user executes an action ($\hat a_\text{user}$ instead of $a_\text{user}$).

\subsection{Intervening on the user's parameters to change user's value function}
\label{sec: user_value_function}
The app agent's reward and transition functions depend on the actions of the user. Therefore it is critical to understand how potential changes to the user's MDP parameters, $\theta$, affect the user's policy. At a given state, the user's policy is to perform the behavior if: 

\begin{align}
    \label{eq: value}
        \underbrace{\sum \limits_{k = 0}^{\delta - 1} \frac{\gamma_\text{user}^k p_\text{user}^k }{z^{k + 1}}\B}_{[1]} + \underbrace{\frac{\gamma_\text{user}^{\delta - 1} p_\text{user}^{\delta} }{z^{\delta}}\G}_{[2]}  
        > \underbrace{\frac{d_\text{world}\D}{1 - \gammau (1-d_\text{world})}}_{[3]}. 
\end{align} 
where $\delta = N - w$ is the user's current distance from the goal state and $z = 1 - \gammau(1 - \p)$. The derivation of \cref{eq: value} is in \cref{appendix: value_derivation}. 

At a high level, component [1] is the burden the user would accumulate in order to reach the goal (and so depends on $\B$), component [2] is the temporally-discounted value of the goal $\G$, and component [3] is the relative consequence of disengagement. From the user's perspective, it is worth trying to reach the goal if the net benefit on the left side outweighs the potential consequence of disengaging on the right side of \cref{eq: value}. Some key insights from this equation that will provide intuition for the intervention strategies we discover in \cref{sec: intervention_strategies}:
\begin{enumerate}[leftmargin=*]
    \item Interventions on $\B$, $\G$, and $\D$ are technically ``unbounded" in effectiveness because they can be any real number. An intervention that makes $\G = 10,000$ could make the inequality hold (thus $a_\text{user}=1$) for any relatively small values of $\B$ and $\D$. 
    \item Interventions on $\gammau$ and $\p$ are ``bounded" in this model because they cannot exceed a value of $1$. Even if an effective app intervention causes $\gamma = 1$ and $\p = 1$, there could still be a situation in which the user is so far from the goal state that they choose not to take action. That is, $(\delta + 1) \B + \G < \D$. 
\end{enumerate}

\section{Insights from our model}
\label{sec: insights}
We are now prepared to user our model to represent different kinds of users and to gain insight on (1) whether different users warrant different intervention strategies, and if so, (2) how these strategies differ. 

\subsection{Representing common user behaviors}
We represent two common decision making impairments-- extreme discounting and mis-estimation of probabilities-- as user agents in our framework. Unless otherwise noted, default parameters for all users in the experiments are: $\gammau = 0.6, \p = 0.6, \G = 10, \B = -1, \D = 0, p_\text{world} = 0.6, d_\text{world} = 0.1$. We also set $\sigma^2 = 0$, meaning that $\hat a_\text{user} = a_\text{user}$. In \cref{appendix: sensitivity}, we include a sensitivity analysis of the empirical results below when the default parameters are sampled from a range. 

\textbf{Users with extreme discounting.}
%A myopic user is one that heavily discounts future rewards. As a result, a myopic user may optimize for immediate gratification, rather than long-term goal achievement. 
The role of extreme temporal discounting in unhealthy behaviors has been explored in smoking cessation, alcohol use, obesity, and many more health applications \citep{story2014discountingAndHealthyBehavoir}. We represent a myopic user with a low discount factor ($\gammau = 0.1$) and a farsighted user with a high discount factor ($\gammau = 0.9$). 

\textbf{Underconfident and overconfident users.}
On certain tasks, human decision-makers tend to report overly-extreme values in estimating probabilities \citep{brenner1996overconfidenceProbability}. In our setting, the user could misestimate the probability of going right. Since, $p_\text{world} = 0.6$ in our experiments, we represent an underconfident user with $\p = 0.1$ and an overconfident user with $\p = 0.9$. 
\begin{figure}[t]
    \centering
     \begin{subfigure}{1\linewidth}
     \centering
     \includegraphics[width=\textwidth]{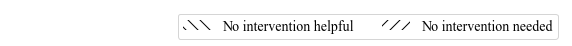}
     \end{subfigure}%   

     \begin{subfigure}{1\linewidth}
     \centering
     \includegraphics[width=\textwidth]{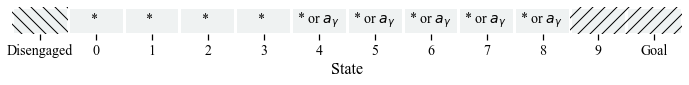}
     \caption{Myopic User}
     \label{fig: myopic-intervention}
     \end{subfigure}
     
     \begin{subfigure}{1\linewidth}
     \centering
     \includegraphics[width=\textwidth]{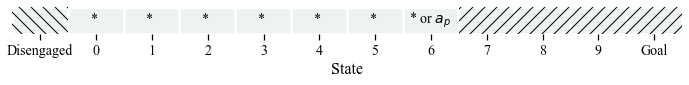}
     \caption{Farsighted User}
     \label{fig: farsighted-intervention}
     \end{subfigure}     
     
     \begin{subfigure}{1\linewidth}
     \centering
     \includegraphics[width=\textwidth]{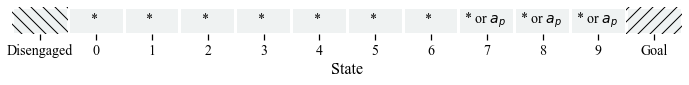}
     \caption{Unconfident User}
     \label{fig: underconfident-intervention}
     \end{subfigure}     
     
     \begin{subfigure}{1\linewidth}
     \centering
     \includegraphics[width=\textwidth]{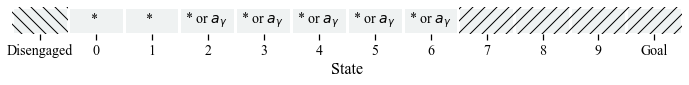}
     \caption{Confident User}
     \label{fig: confident-intervention}
     \end{subfigure}
    \caption{\textbf{The app agent's intervention strategy differs for different types of users}. App's policy is plotted. The $*$ represents ``$a_\B$ or $a_\D$."  In some states,
    no intervention could make the user move (when $w = \text{disengaged}$). In other states, the user's default policy is to move, even without intervention (when $w=7$ in \cref{fig: confident-intervention}).}
    \label{fig: main policies}
\end{figure}%

\subsection{Intervention strategies}
\label{sec: intervention_strategies}
%In this section, we explore (1) what intervention to use and when and (2) how effective each intervention must be for different types of users. 

%Before discussing the interventions, we will begin with a brief, demonstrative example of an interaction between the user and app agent using the results from \cref{fig: main policies}. Suppose the app is interacting with a myopic user who is in state $w = 4$ (\cref{fig: myopic-intervention}). Without intervention, the user's optimal policy under default parameters is not to do PT. Here, the app's policy is to intervene on $\B$ or $\D$. Suppose the app intervenes on $\B$, so that, in the moment, the user's policy under the updated parameters is to do PT, which moves them into state $w = 5$. Once $w = 5$, the app may intervene on $\p$, $\B$, or $\D$. 

\begin{figure}[t]
    \centering
     \begin{subfigure}{0.4\linewidth}
     \centering
     \includegraphics[width=\linewidth]{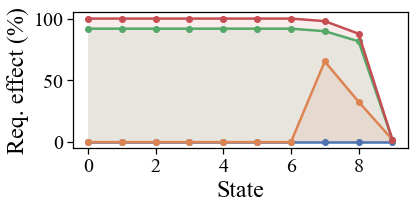}
     \caption{Underconfident}
     \label{fig: underconfident-effect}
     \end{subfigure}%
     \begin{subfigure}{0.4\linewidth}
     \centering
     \includegraphics[width=\linewidth]{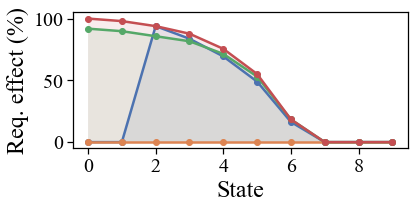}
     \caption{Overconfident}
     \label{fig: overconfident-effect}
     \end{subfigure}
     \begin{subfigure}{0.1\linewidth}
     \includegraphics[width=\linewidth]{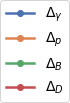}%
     \end{subfigure}
     \caption{\textbf{Within a user type, some parameters may be easier to intervene on than others}. Lines represent the minimum effectiveness $\left ( \frac{\Delta}{\max(\Delta)} \times 100\right)$ the intervention must have in order to change the user's behavior, where $\max(\Delta)$ is the maximum feasible effect on a parameter. For example, $\B + \max(\Delta_B) > 0$ and $\gammau + \max(\Delta_\gamma) = 1$. Lower is better; low means the intervention can have a small $\Delta$ but still change behavior. Remaining users in \cref{appendix: required_interventions}.}
    \label{fig: effect_size}
\end{figure}%

\subsubsection{What interventions to use and when?}
% Talk about the existence of windows. 
Inspection of \cref{fig: main policies} reveals a recurring pattern in the intervention strategies. The app's policy always contains two ``windows": \textbf{Window 1} covers states that are far from the reward and consists of interventions on $\B$ or $\D$. \textbf{Window 2} covers states that are closer to the reward and consists of interventions \emph{only} on $\p$ or \emph{only} on $\gammau$, depending on the type of user. Below, the intuition for why such windows exist. 

\emph{\textbf{Window 1: interventions on $\B$ or $\D$.}}
%Across all user types in \cref{fig: main policies}, the intervention strategy in Window 1 is to intervene on $\B$ or $\D$.  
Interventions in this window target user parameters that can affect user behavior regardless of distance from the goal state. Mathematically, this is due to the observation we made about the unboundedness of the value of $\B$, $\D$, and $\G$ in \cref{sec: user_value_function}. Intuitively, if we can make the user feel enjoyment during PT, then they will work out even if it does not help them move toward the goal. In our model, this corresponds to an intervention that makes $B$ positive. If we make the consequence of disengagement very high, such as a reminder that the consequence of quitting PT is to never regain full mobility of the shoulder, then the user will work out. In our model, this corresponds to an intervention which makes the right hand threshold in $\cref{eq: value}$ very negative. 

\emph{\textbf{Window 2: depending on user type, interventions on $\p$ or $\gammau$.}} For myopic and overconfident users, Window 2 interventions are on $\boldsymbol{\gammau}$. For underconfident and farsighted users, Window 2 interventions are on $\boldsymbol{\p}$. The respective interventions on $\gammau$ and $\p$ make sense for the myopic and underconfident users, since these are the parameters that are directly impairing the decision making. For overconfident users, intuition comes from realizing that when $\p \approx 1$ in \cref{eq: value}, the relative values of $\gamma, \B, \G, \D$ affect the decision making. Since $\B$ and $\D$ are considered in Window 1, this leaves $\gammau$ in Window 2. The same holds by considering $\gammau \approx 1$ for farsighted users. 

\textbf{For each user, the \emph{best} intervention depends on the effectiveness of the intervention.}
%We refer to an intervention as ``valid" for a given state $w$ if it would make the user take action assuming it had the maximum intended effect. \sam{prior sentence is unclear.  maybe break into 2 sentences? Anyway I am not really sure what prior sentence means.  }  \sam{in the phrase "We refer to an intervention as ``valid" in a given state"  are we referring to user state or app stat?  Would be good to make clear.} For interventions on $\B$, a valid  intervention  makes $\B > 0$. For interventions on $\gammau$, a valid intervention  makes $\gammau = 1$. \sam{Is an intervention that makes $p_{user}=1$ always valid? --I guess not.  might point this out to reader.}

All of the policies shown in \cref{fig: main policies} assume that the interventions are maximally effective. In reality, the effect of an intervention may vary across users, even within the same user type. For example, one underconfident user may respond well to interventions on $D$ (i.e. $\Delta_D$ is large), while another may not be impacted by such interventions at all (i.e. $\Delta_D = 0$). Consider \cref{fig: underconfident-effect}. Suppose the underconfident user is in state $w = 7$. Although intervening on $\p$, $\B$, or $\D$ is valid, the interventions on $\B$ and $\D$ would require a high level of effectiveness near $100\%$, while the intervention on $\p$ must be at least $75\%$ effective. To account for variability in effectiveness for different underconfident users, it may be best to intervene on $\p$ in this scenario. However, for the overconfident user in \cref{fig: overconfident-effect}, all interventions truly seem equally preferable, since they require similar levels of effectiveness. 

\section{Conclusion and Future Work}
In this work, we represented the user's decision making process with reinforcement learning and formalized how interventions from the app agent effect the user's \emph{MDP parameters}. 
%We implemented four types of user agents-- a myopic, farsighted, overconfident, and underconfident user--and explored how the app agent's intervention strategy differs for each user type. 
Our framework exposes interesting research questions that across computer and behavioral science. 

\textbf{Computational questions.} Do these computational insights generalize to real world users? Assuming we can infer a user's parameters when they download the mHealth app, how do we incorporate knowledge of these parameters in guiding the app's interventions? How do we modify this framework to represent interventions that have a permanent, or at least a multi-timestep effect, on the user's parameters? 

\textbf{Behavioral} What are the most worthwhile ways to add complexity to the world model in order to make it more representative of human behavior (e.g. making probability of disengagement depend on state, adding more states, etc.)? What set of questions, cognitive tasks, or sensor readings can we use to estimate the user's MDP parameters? How do we design effective interventions for each of these MDP parameters? 

\section{Acknowledgements}
This material is based upon work supported by the National Science Foundation under Grant No. IIS-2107391 and the National Institute of Biomedical Imaging and Bioengineering and the Office of the Director of the National Institutes of Health under award number P41EB028242.  Any opinions, findings, and conclusions or recommendations expressed in this material are those of the author(s) and do not necessarily reflect the views of the National Science Foundation.

\bibliography{main}

\appendix
\onecolumn
\section{Derivation of the user's optimal value function}
\label{appendix: value_derivation}
\newcommand{\vstay}{{V^{\pi_\text{stay}}}}
\newcommand{\vright}{{V^{\pi_\text{right}}}}
\newcommand{\given}{{\ \middle | \ }}
In this section, we derive the value function of the user's optimal policy, $V_\text{user}^{\pi^*}$, with respect to a user's MDP parameters. For brevity, we will refer to this as $V^*$. Similarly, while the user's MDP parameters are subscripted in the main text (e.g. $B_\text{user}$), we will forgo these subscripts here ($B$).

In our setting, once the optimal policy is to go right in a given state, the best strategy is to continue going right in subsequent states that are closer to the goal. That is, if $\pi^*(w) = 1$, then $\pi^*(w + 1) = 1$. The opposite is also true; if the optimal policy is to stay in place in a given state, then the best strategy in a state that is farther away from the goal is also to stay in place --- if $\pi^*(w) = 0$, then $\pi^*(w - 1) = 0$. 

The optimal policy chooses the maximum value between the value of a policy that always chooses to go right, $\pi_\text{right}(w) = 1$, and the value of a policy that always chooses to stay in place, $\pi_\text{stay}(w) = 0$:

\begin{equation}
\label{eq: generic opt value func}
    V^*(w) = \max\{\vstay(w), \vright(w)\}
\end{equation}

The derivation for $\vstay(w)$ is as follows: 
\begin{align}
    & \vstay(w) \\
    & = \mathbb{E}\left [\sum\limits_{t = 0}^\infty \gamma^t R_t \right ]\\
    & = \underbrace{\gamma^0 \left [d^1 D + (1 - d)^1 0 \right]}_{R_0} + \underbrace{(1 - d)^1 \gamma^1 \left [d D + (1 - d) 0 \right]}_{R_1} + \underbrace{(1 -d)^2 \gamma^2 \left [d D + (1 - d) 0 \right]}_{R_2} + \ldots\\
    & = \sum\limits_{t = 0}^\infty \gamma^t (1-d)^{t} d D\\
    & = \frac{d D}{1 - \gamma (1-d)}
\end{align}

We will derive $\vright$ for specific states, and generalize these findings to a general equation. First, we will derive the value of a state which is right before the goal state, $w = N - 1$: 

\begin{align}
\begin{split}
\label{eq: v1}
    & \vright(N - 1) \\
    & = \mathbb{E}\left [\sum\limits_{t = 0}^\infty \gamma^t R_t \right ]\\
    & = \underbrace{pG + B}_{\text{ when } s_0 = N - 1}+ \underbrace{\gamma (1 - p)[pG + B]}_{\text{ when } s_1 = N - 1 }+ \underbrace{\gamma^2 (1 - p)^2[pG + B]}_{\text{ when } s_2 = N - 1} + \ldots\\
    & = \sum\limits_{t = 0}^\infty \gamma^t (1 - p)^t [pG + B]\\
    & = \frac{pG + B}{1 - \gamma(1 - p)}\\
    & = \frac{pG + B}{z}.
\end{split}
\end{align}

In \cref{eq: v1}, the $pG + B$ term represents the expected reward of taking action $a = 1$ in state $w = N - 1$. Recall that when $a=1$ the user moves right with probability $p$ and stays in place with probability $1 - p$. The user always receives a reward of $B$ for choosing $a=1$ and potentially a reward of $G$ if they transition to the goal state. Then the calculation for the expected reward is: 

\begin{align}
    \begin{split}
        & \mathbb{E}[R_t |  s_t = N - 1,\ a_t = 1]\\
        & = p (G + B) + (1 - p) B\\
        & = pG + B\\
    \end{split}
\end{align}

Next, we derive the value of a state which is two spaces away from the goal state, $w = N - 2$: 
\begin{align}
\begin{split}
    \label{eq: v2}
    & \vright(N - 2) \\
    & = r(w = N - 2, a = 1) + \gamma \sum\limits_{w'} P(w' | w = N-2, a = 1) \vright(w')\\
    & = B + \gamma \left [ p \vright(N - 1) + (1 - p) \vright(N-2) \right ] \\
    & = B + \gamma  p \vright(N - 1) + \gamma (1 - p) \vright(N-2)  \\
    & = \underbrace{ B + \gamma  p \vright(N - 1)}_{\text{ when } s_0 = N - 2} 
    + \underbrace{\gamma (1 - p) [ B + \gamma  p \vright(N - 1)]}_{\text{ when } s_1 = N - 2 } 
    + \underbrace{\gamma^2 (1 - p)^2 [ B + \gamma  p \vright(N - 1)]}_{\text{ when } s_2 = N - 2 } + \ldots\\
    & = \sum\limits_{t = 0}^\infty \gamma^t(1-p)^t [ B + \gamma  p \vright(N - 1)]\\
    & = \frac{ B + \gamma  p \vright(N - 1)}{1 - \gamma(1 - p)}\\
    & = \frac{ B + \gamma  p \vright(N - 1)}{z}\\
    & = \frac{B + \gamma p \frac{pG + B}{z}}{z}\\
    & = \frac{\gamma p^2 G}{z^2} + \frac{\gamma p B}{z^2} + \frac{B}{z}.\\
\end{split}
\end{align}

Notice that in \cref{eq: v2}, we can apply the Bellman equation to ``recursively" expand the form of the value function, so that it can be written as an infinite geometric series: 

\begin{align}
\label{eq: recursive_value}
    \vright(w) = \sum \limits_{t = 0}^\infty \gamma^t (1-p)^t \left [ B + \gamma p \vright(w + 1) \right ]
\end{align}

We will apply \cref{eq: recursive_value} to our final derivation of $\vright(N - 3)$:
\begin{align}
\begin{split}
    & \vright(N - 3) \\
    & = \sum\limits_{t = 0}^\infty \gamma^t(1-p)^t [B + \gamma p \vright(N-2) ]\\
    & = \frac{B + \gamma p \vright(N-2) }{1 - \gamma(1 - p)}\\
    & = \frac{B + \gamma p \vright(N-2)}{z}\\
    & = \frac{B + \gamma p \left(\frac{\gamma p^2 G}{z^2} + \frac{\gamma p B}{z^2} + \frac{B}{z}\right)}{z}\\
    & = \frac{\gamma^2 p^3}{z^3}G + \frac{\gamma^2 p^2}{z^3}B + \frac{\gamma p}{z^2} + \frac{B}{z}.\\
\end{split}
\end{align}

In general, for any state $w = N - \delta$, the value function is: 
\begin{equation}
    \vright(w = N - \delta) = \frac{\gamma^{\delta - 1} p^\delta }{z^\delta}G + \sum\limits_{k = 0}^{\delta - 1}\frac{\gamma^k p^k}{z^{k + 1}}B,
\end{equation} 
where $z = 1 - \gamma(1 - p)$. 

Recall \cref{eq: generic opt value func}, where we defined the optimal value function with respect to $\vright$ and $\vstay$. Replacing these terms with their analytical counterparts, we arrive at: 
\begin{equation}
    V^*(w = N - \delta) = \max \left \{\underbrace{\frac{dD}{1 - \gamma (1-d)}}_{\vstay},  \underbrace{\frac{\gamma^{\delta - 1} p^\delta}{z^\delta}G + \sum\limits_{k = 0}^{\delta - 1}\frac{\gamma^k p^k}{z^{k + 1}}B}_\vright \right \} 
\end{equation}

Along the same lines, the user's optimal policy is to follow the action of the policy that maximizes the value in the current state: 
\begin{equation}
    \pi^*(w = N - \delta) = \mathbb{I} \left \{\underbrace{\frac{ \gamma^{\delta - 1} p^\delta}{z^\delta}G + \sum\limits_{k = 0}^{\delta - 1}\frac{\gamma^k p^k}{z^{k + 1}}B}_\vright > \underbrace{\frac{dD}{1 - \gamma (1-d)}}_{\vstay} \right \} 
\end{equation}
\section{Required intervention effect for all user types}
\label{appendix: required_interventions}
\begin{figure}[h]
\centering
 \begin{subfigure}{0.25\linewidth}
    \centering
    \includegraphics[width=\linewidth]{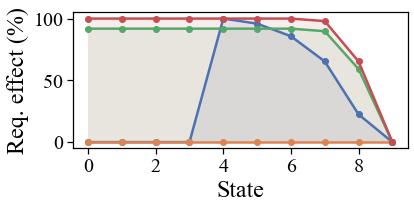}
    \caption{Myopic users}
    \label{}
\end{subfigure}%
 \begin{subfigure}{0.25\linewidth}
    \centering
    \includegraphics[width=\linewidth]{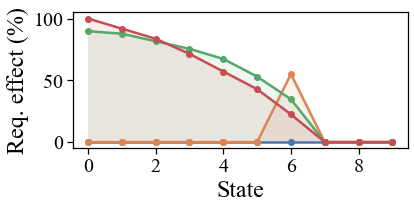}
    \caption{Farsighted users}
    \label{}
\end{subfigure}%
 \begin{subfigure}{0.25\linewidth}
    \centering
    \includegraphics[width=\linewidth]{figures/Underconfident_user-effect.png}
    \caption{Uunderconfident users}
    \label{}
\end{subfigure}%
 \begin{subfigure}{0.25\linewidth}
    \centering
    \includegraphics[width=\linewidth]{figures/Overconfident_user-effect.png}
    \caption{Overconfident users}
    \label{}
\end{subfigure}%
\caption{\textbf{Within a user type, some parameters may be easier to intervene on than others}. Lines represent the minimum effect $\left ( \frac{\Delta}{\max(\Delta)} \times 100\right)$ the intervention must have in order to change the user's behavior, where $\max(\Delta)$ is the maximum feasible effect on a parameter. For example, $\B + \max(\Delta_B) > 0$ and $\gammau + \max(\Delta_\gamma) = 1$. Lower is better; it means the intervention can have a small effect but still achieve the same behavior change.}
\end{figure}

\section{Sensitivity analysis: sampling the default parameters}
\label{appendix: sensitivity}
The ``default" parameters for the user MDP from the main experiments are recapped in the first column of \cref{table: default_parameters}. We checked that the main results of the paper are not sensitive to these parameter settings. To do so, we generated multiple random samples of different default parameters and confirmed that the main intervention patterns from the results still hold. To form the default parameter set, each parameter is uniformly sampled from the ranges in the second column of \cref{table: default_parameters}. 

We kept the values of $p_\text{world}$ and $G$ fixed in this analysis. We fix $p_\text{world} = 0.6$ because the app cannot intervene on $p_\text{world}$ and the user's policy does not depend on $p_\text{world}$, and we do not expect $p_\text{world}$ will not affect the app's policy. We fix the value of $\G = 10$ because we do not consider interventions on $\G$ and because the value of $\G$ is meaningful only in relation to the value of $\B$, which we \emph{do} sample.

\begin{table}[h]
    \centering
    \begin{tabular}{c|c}
         \textbf{Default parameters in main body} & \textbf{Sampled parameter range} \\
         $\gammau = 0.6$ &  $\gammau \in [0.4, 0.7]$\\
         $\p = 0.6$ & $\p \in [0.5, 0.7]$\\
         $\B = -1$ & $\B \in [-5, -0.1]$ \\
         $\D = 0$ & $\D \in [-5, 5]$\\
         $d_\text{world} = 0.1$ & $d_\text{world} \in [0.1, 0.5]$\\
         $\G = 10$ & not sampled \\
         $p_\text{world} = 0.1$ & not sampled \\
    \end{tabular}
    \caption{Default user MDP parameters used in the experiments in the main body (left) and the range from which they are sampled in the sensitivity analysis (right).}
    \label{table: default_parameters}
\end{table}

The main patterns that we expect will continue to hold across different samples of default parameters are: 
\begin{enumerate}
    \item The myopic and overconfident users will require intervention on $\B$ or $\D$ in Window 1 and then intervention on $\gamma$ in Window 2 (the sizes of these windows will vary). 
    \item The underconfident and farsighted users will require intervention on $\B$ or $\D$ in Window 1 and then intervention on $\p$ in Window 2 (the sizes of these windows will vary). 
    \item Right before the goal (i.e. in Window 3), it may be the case that an intervention on any parameter makes enough of a difference to nudge the user to take action. 
\end{enumerate}

These patterns hold across $5$ different samples of the default parameters, shown in REF.

\begin{figure}[h]
    \centering
    \begin{subfigure}{0.5\linewidth}
    \centering
    \includegraphics[width=\linewidth]{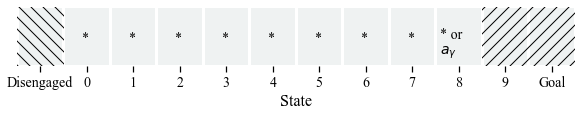}
    \caption{App policy for \textbf{myopic} user}
    \label{}
    \end{subfigure}%
    \begin{subfigure}{0.5\linewidth}
    \centering
    \includegraphics[width=\linewidth]{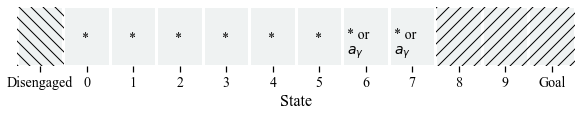}
    \caption{App policy for \textbf{overconfident} user}
    \label{}
    \end{subfigure}%
    
    \begin{subfigure}{0.5\linewidth}
    \centering
    \includegraphics[width=\linewidth]{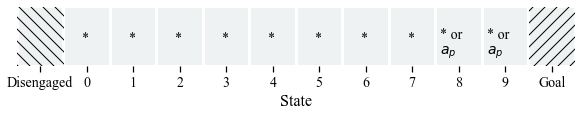}
    \caption{App policy for \textbf{underconfident} user}
    \label{}
    \end{subfigure}%
    \begin{subfigure}{0.5\linewidth}
    \centering
    \includegraphics[width=\linewidth]{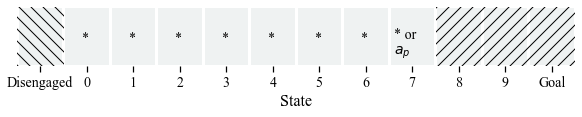}
    \caption{App policy for \textbf{farsighted} user}
    \label{}
    \end{subfigure}%
    \caption{\textbf{The expected results hold for trial 1} with sampled parameters $\p = 0.61, \d = 0.39, \gammau = 0.58, \B = -2.33, \D = -0.76$}
    \label{fig: sensitivity_trial_1}
\end{figure}%

\begin{figure}[h]
    \centering
    \begin{subfigure}{0.5\linewidth}
    \centering
    \includegraphics[width=\linewidth]{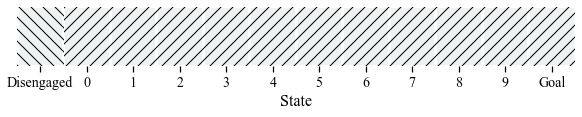}
    \caption{App policy for \textbf{myopic} user}
    \label{}
    \end{subfigure}%
    \begin{subfigure}{0.5\linewidth}
    \centering
    \includegraphics[width=\linewidth]{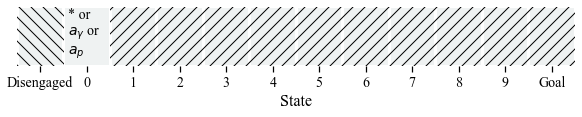}
    \caption{App policy for \textbf{overconfident} user}
    \label{}
    \end{subfigure}%
    
    \begin{subfigure}{0.5\linewidth}
    \centering
    \includegraphics[width=\linewidth]{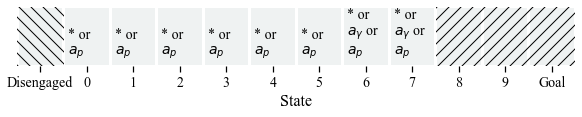}
    \caption{App policy for \textbf{underconfident} user}
    \label{}
    \end{subfigure}%
    \begin{subfigure}{0.5\linewidth}
    \centering
    \includegraphics[width=\linewidth]{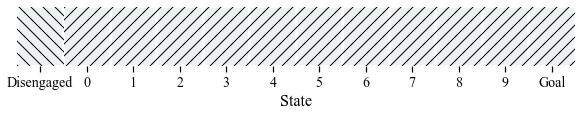}
    \caption{App policy for \textbf{farsighted} user}
    \label{}
    \end{subfigure}%
    \caption{\textbf{The expected results hold for trial 2} with sampled parameters $\p = 0.63, \dreal = 0.28, \gammau = 0.67, \B = -0.28, \D = -1.17$}
    \label{fig: sensitivity_trial_2}
\end{figure}%

\begin{figure}[h]
    \centering
    \begin{subfigure}{0.5\linewidth}
    \centering
    \includegraphics[width=\linewidth]{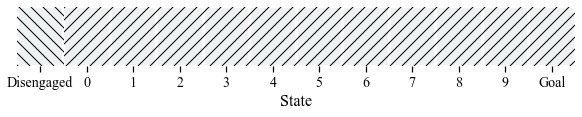}
    \caption{App policy for \textbf{myopic} user}
    \label{}
    \end{subfigure}%
    \begin{subfigure}{0.5\linewidth}
    \centering
    \includegraphics[width=\linewidth]{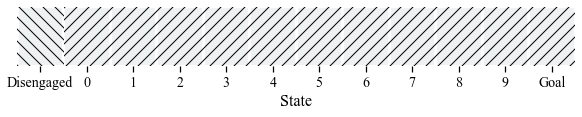}
    \caption{App policy for \textbf{overconfident} user}
    \label{}
    \end{subfigure}%
    
    \begin{subfigure}{0.5\linewidth}
    \centering
    \includegraphics[width=\linewidth]{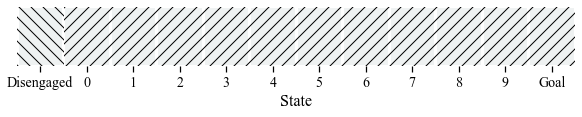}
    \caption{App policy for \textbf{underconfident} user}
    \label{}
    \end{subfigure}%
    \begin{subfigure}{0.5\linewidth}
    \centering
    \includegraphics[width=\linewidth]{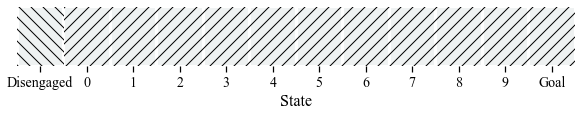}
    \caption{App policy for \textbf{farsighted} user}
    \label{}
    \end{subfigure}%
    \caption{\textbf{For trial 3, all user's policies are to take action by default.} This is likely due to the small value of $\B$. The parameters are: $\p = 0.66, \dreal = 0.31, \gamma = 0.57, \B = -0.46, \D = -4.29$}
    \label{fig: sensitivity_trial_3}
\end{figure}%

\begin{figure}[h]
    \centering
    \begin{subfigure}{0.5\linewidth}
    \centering
    \includegraphics[width=\linewidth]{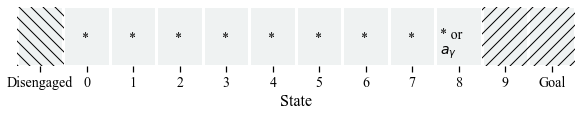}
    \caption{App policy for \textbf{myopic} user}
    \label{}
    \end{subfigure}%
    \begin{subfigure}{0.5\linewidth}
    \centering
    \includegraphics[width=\linewidth]{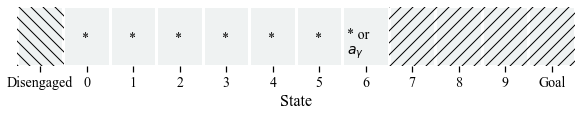}
    \caption{App policy for \textbf{overconfident} user}
    \label{}
    \end{subfigure}%
    
    \begin{subfigure}{0.5\linewidth}
    \centering
    \includegraphics[width=\linewidth]{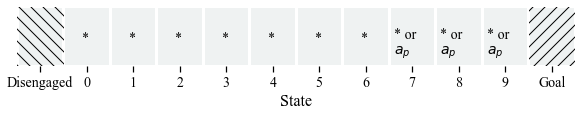}
    \caption{App policy for \textbf{underconfident} user}
    \label{}
    \end{subfigure}%
    \begin{subfigure}{0.5\linewidth}
    \centering
    \includegraphics[width=\linewidth]{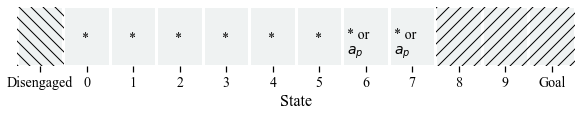}
    \caption{App policy for \textbf{farsighted} user}
    \label{}
    \end{subfigure}%
    \caption{\textbf{The expected results hold for trial 4} with sampled parameters $\p = 0.52, \dreal = 0.11, \gammau = 0.65, \B = -1.19, \D = 3.70$}
    \label{fig: sensitivity_trial_4}
\end{figure}%

\begin{figure}[h]
    \centering
    \begin{subfigure}{0.5\linewidth}
    \centering
    \includegraphics[width=\linewidth]{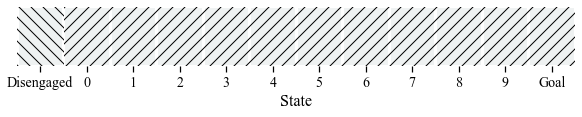}
    \caption{App policy for \textbf{myopic} user}
    \label{}
    \end{subfigure}%
    \begin{subfigure}{0.5\linewidth}
    \centering
    \includegraphics[width=\linewidth]{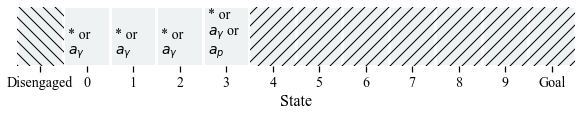}
    \caption{App policy for \textbf{overconfident} user}
    \label{}
    \end{subfigure}%
    
    \begin{subfigure}{0.5\linewidth}
    \centering
    \includegraphics[width=\linewidth]{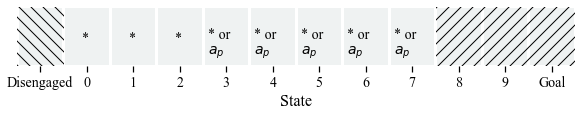}
    \caption{App policy for \textbf{underconfident} user}
    \label{}
    \end{subfigure}%
    \begin{subfigure}{0.5\linewidth}
    \centering
    \includegraphics[width=\linewidth]{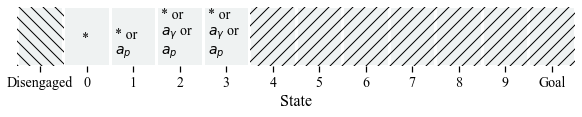}
    \caption{App policy for \textbf{farsighted} user}
    \label{}
    \end{subfigure}%
    \caption{\textbf{The expected results hold for trial 5} with sampled parameters $\p = 0.70, \dreal = 0.42, \gammau = 0.54, \B = -1.18, \D = -3.82$}
    \label{}
    \label{fig: sensitivity_trial_5}
\end{figure}%
\end{document}